\documentclass[conference,letterpaper,twoside,final,10pt]{ieeeconf}
\usepackage[left=1.91cm,right=1.91cm,top=1.91cm,bottom=1.91cm]{geometry}

\usepackage{graphicx}
\usepackage{amssymb}
\usepackage{amsmath}
\usepackage{cite}
\usepackage{comment}
\usepackage{color}
\usepackage{url}
\usepackage{alltt}
\usepackage{cleveref}
\usepackage{tikz}
\usepackage{pgfplots}
\usepackage{pgf-umlsd}
\usepackage{cprotect}
\usepackage{multirow}
\usepackage{algorithm}
\usepackage[noend]{algpseudocode}
\usepackage{bigfoot}
\usepackage{paralist}
\usepackage{tabularx}
\usepackage[whole]{bxcjkjatype}
\usepackage{ifthen}
\usepackage{threeparttable}
\usepackage[hang,small,bf]{caption}
\usepackage[subrefformat=parens]{subcaption}
\usepackage[super]{nth}
\usepackage{colortbl}
\usepackage{scalefnt}
\usepackage{threeparttable}

\pgfplotsset{compat=1.14}
\IEEEoverridecommandlockouts

\makeatletter
\newcommand{\figcaption}[1]{\def\@captype{figure}\caption{#1}}
\newcommand{\tblcaption}[1]{\def\@captype{table}\caption{#1}}
\makeatother

\title{\LARGE \bf
Uncertainty-aware Self-supervised Target-mass Grasping\\
of Granular Foods
}

\author{
Kuniyuki Takahashi$^{\dagger}$,
Wilson Ko$^{\dagger}$,
Avinash Ummadisingu$^{\dagger}$,
Shin-ichi Maeda$^{\dagger}$
\thanks{$^{\dagger}$ K. Takahashi, W, Ko,  A. Ummadisingu, and S. Maeda are associated with Preferred Networks, Inc.
\ takahashi@preferred.jp \ wko@preferred.jp \ ummavi@preferred.jp \ ichi@preferred.jp}}

\begin{document}

\maketitle
\thispagestyle{empty}

\begin{abstract}
Food packing industry workers typically pick a target amount of food by hand from a food tray and place them in containers.
Since menus are diverse and change frequently, robots must adapt and learn to handle new foods in a short time-span.
Learning to grasp a specific amount of granular food requires a large training dataset, which is challenging to collect reasonably quickly.
In this study, we propose ways to reduce the necessary amount of training data by augmenting a deep neural network with models that estimate its uncertainty through self-supervised learning.
To further reduce human effort, we devise a data collection system that automatically generates labels.
We build on the idea that we can grasp sufficiently well if there is at least one low-uncertainty (high-confidence) grasp point among the various grasp point candidates.
We evaluate the methods we propose in this work on a variety of granular foods- coffee beans, rice, oatmeal and peanuts, each of which has a different size, shape and material properties such as volumetric mass density or friction.
For these foods, we show significantly improved grasp accuracy of user-specified target masses using smaller datasets by incorporating uncertainty.\footnote{An accompanying video is available at the following link:\\ \url{https://youtu.be/5pLkg7SpmiE}}

\end{abstract}
\section{Introduction}
\label{sec:introduction}
Within the food packing industry, the ability to manipulate multiple, highly variable food and place them in their assigned positions is essential.
This requirement is exemplified by the Japanese lunch boxes known as ``Bentos''.
Bentos often combine numerous types of food and change in contents every few weeks to make use of seasonal ingredients.
Not only is there immense inter-category variation in the types of food that need to be manipulated, but there is also significant intra-category irregularity in shape, size.
The continually changing menus and variation make automation of the process a significant challenge and has so far only been applied to a small portion of mostly fixed foodstuffs.
In times of labor shortages, it has become crucial to develop intelligent systems that can learn to adapt to this industry's variety, pace and scale.

\begin{figure}[tbp]
	\centering
	\includegraphics[width=0.80\columnwidth]{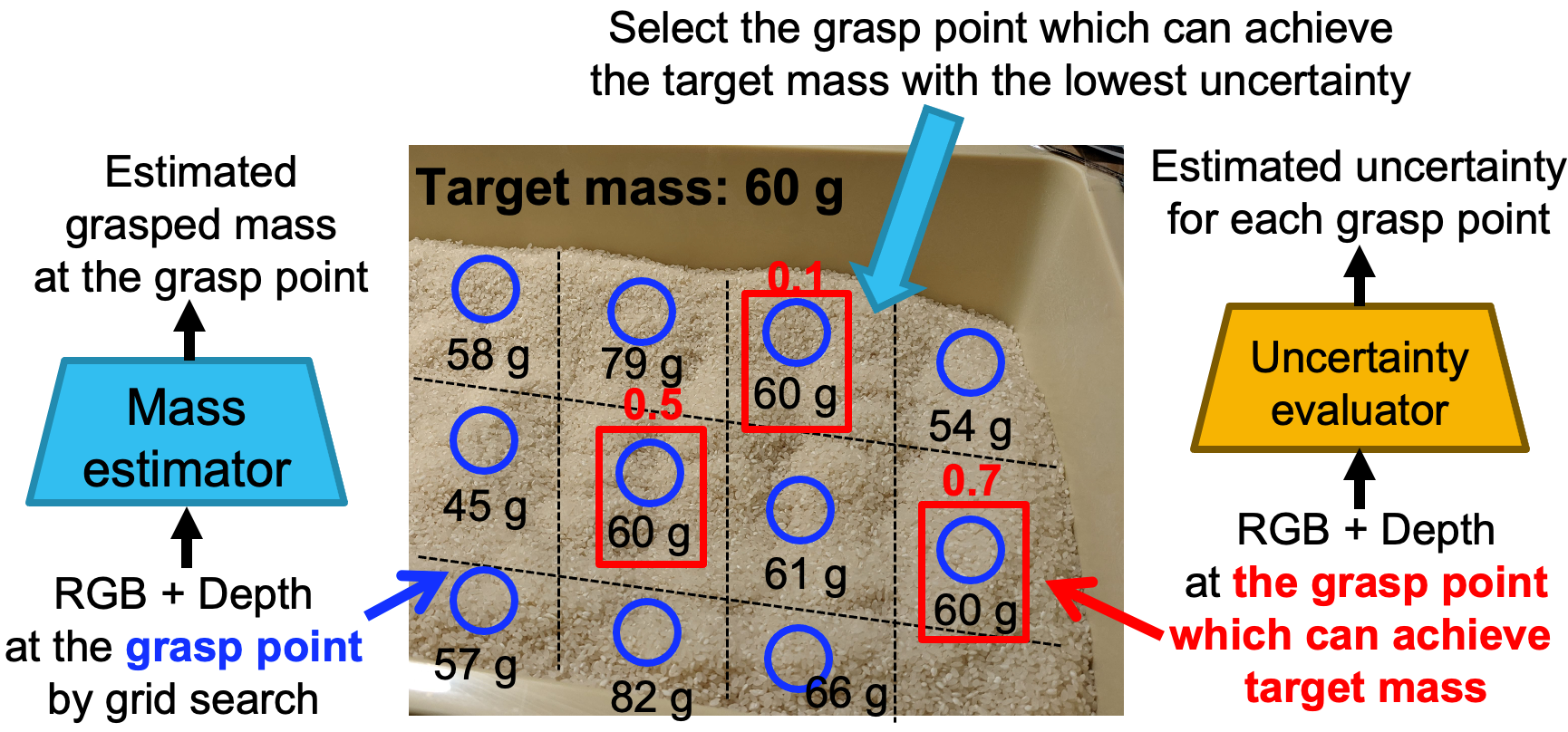}
	\caption{Concept: The robot estimates grasp masses at several grasp points. The uncertainty of the prediction is evaluated at the grasp points which are close to the target mass. Then, the grasp point sufficiently close to the target mass with the lowest uncertainty (highest confidence) is selected.}
	\label{fig:concept}
\end{figure}

The amount of food to be placed in a bento box is specified either by the number of pieces for \emph{solid food} (e.g. fried chicken and fish) or by mass for \emph{granular food} (e.g. beans and shredded vegetables).
Prior work in robotics has been successfully demonstrated for the manipulation of \emph{solid foods} in the food industry~\cite{matsuoka2020learning, misimi2018robotic}.
However, despite a substantial number of used \emph{granular foods}, their manipulation remains relatively unexplored and they are difficult to deal with due to their particulate, deformable nature.
So far, specialized tools or measured scoops are often deployed to measure the required quantity of food, but this often requires significant engineering effort to adapt to different foods and amounts.

In our work, we focus on learning to pick \emph{granular food} by identifying the best pick points to grasp a user-specific target mass from an RGB-D image of the entire food tray (See Fig.~\ref{fig:concept}).
We use feed-forward planning based on vision in order to avoid the cost of additional sensors, such as wrist-based load cells.
In designing this system, there are three requirements from the food packing industry.
R1) The ability to handle a variety of deformable and irregularly shaped food.
R2) The ability to grasp any user-defined target mass of food without the need for additional training.
R3) The ability to adapt to new food in a short period of time.
We propose a grasping method that satisfies these three requirements.

Since granular foods are deformable and the individual particles heavily interact with each other, it is difficult to simulate accurately or model. 
A learning-based approach from real-world interaction was hence the most feasible choice for our work.
Our proposed system leverages the representational power of deep learning to learn from real-world RGB-D images directly (R1).
The aim of the neural network is to predict the mass of grasped food for several candidates within the tray.
The most likely grasp to result in the user-requested mass is chosen and executed (R2).
Although deep learning is powerful, it usually needs a large training dataset to be optimized correctly.
Unfortunately, data-collection on the real robot takes an enormous amount of time and working with real food is challenging due to spoilage, damage due to handling, moisture, etc, and thus limits the amount of data we can collect and learn from.
This limitation inevitably makes the estimate of the trained deep neural networks unreliable.
To cope with the incompleteness, we model the uncertainty of the estimate with self-supervised learning and utilize the most reliable estimate (R3).

\section{Related Work}
\label{sec:related works}
\subsection{Granular Food Manipulation}
\label{sec:Granular Food Manipulation}
For granular food manipulation, prior work is typically limited to a specific food or task, e.g. in~\cite{hoerger2019pomdp}, the authors created a candy-scooping robot.
In~\cite{schenck2017learning}, the robot learned a scooping motion of pinto beans from one tray to another to arrange them into the desired shape.
A study by~\cite{clarke2018learning} is highly relevant to our research.
The system proposed in this work scoops a user-specified target mass of food using a box-shaped end-effector, given the depth information of the food tray, where it tries to pick up food with a single grasp position with fixed orientation.
In scooping, the insertion position, angle and depth of the gripper are required at minimum (such as in~\cite{clarke2018learning}) in order to control the trajectory.
On the other hand, gripping only requires the position and insertion depth of the gripper.
Therefore, grasping requires fewer parameters to be executed.
We chose grasping with a gripper over scooping to simplify the picking process so that we can focus on reducing the required training data.
Additionally, we improve inference performance and further reduce training data by taking uncertainty into account.
Note that our methods of uncertainty estimation can also be applied for scooping.

\subsection{Estimation of Uncertainties}
\label{sec:related Estimation of Uncertainties}
Literature on uncertainty makes a distinction between two sources of uncertainty- \emph{epistemic} and \emph{aleatoric}~\cite{hora1996aleatory}.
Epistemic (or systematic) uncertainty stems from the lack of knowledge or from not knowing or accounting for results outside of what is encountered so far.
Epistemic uncertainty is considered reducible, as the acquisition of more data or knowledge of the data generating process would reduce it.
Aleatoric (or statistical) uncertainty represents the inherent randomness or variability in the outcome of an experiment.
It would typically cover uncertainty in sensing, the robot dynamics, and the future state of the environment after interaction.

A number of recent works aim at introducing uncertainty into neural networks.
From the Bayesian perspective, aleatoric uncertainty may be modelled by placing a distribution over the output of the network~\cite{kendall2017uncertainties} and epistemic uncertainty is modelled by placing a prior distribution over a model's weights.
The latter's formulation is often referred to as Bayesian Neural Networks to~\cite{mackay1992bayesian, Buntine:91, neal1995, freitas2003, gal2016uncertainty, denker1991transforming} and their related extensions~\cite{gal2015bayesian,gal2016uncertainty,kendall2017uncertainties, gal2015dropout}.
However, their adoption, as well as methods that employ ensembles~\cite{lakshminarayanan2017simple} that capture both kinds of uncertainty, are computationally expensive.

As we operate in the small-data regime, we adopt methods from ML literature to explicitly capture uncertainty and use it in a two-stage grasp selection criterion that chooses the least uncertain grasp that leads to our target mass.

\subsection{Grasping}
\label{sec:related Grasping}
Grasp synthesis, or the problem of choosing a grasp configuration among an infinite set of candidates has been addressed frequently in the robotics community~\cite{bicchi2000robotic, bohg2013data, sahbani2012overview}.
Prior work is broadly classified into \emph{analytical methods} that explicitly model and/or deal with one or more sources of (aleatoric) uncertainty~\cite{brost1988automatic, goldberg1990bayesian, zheng2005coping, christopoulos2007handling} or \emph{data-driven methods}~\cite{goldfeder2011data,bohg2013data} that grew with the advent of Graspit!~\cite{miller2004graspit} which allow for a less strict parametrization of the grasp and therefore accommodate better for uncertainties in perception and execution~\cite{goldfeder2011data, hsiao2009reactive, pastor2011online, hsiao2010contact, kim2013physically}. 
Applications of neural networks to grasp synthesis tend to use adapted versions of popular CNN architectures to output the grasp quality metric for either a single grasp, a collection of grasp candidates, or directly output the best grasp candidate.
They are typically trained on synthetic data~\cite{johns2016deep, mahler2017dex, miller2003automatic}, using datasets like the Cornell object dataset~\cite{lenz2015deep} in~\cite{lenz2015deep, redmon2015real, kumra2017robotic, morrison2018closing} or learn from sizable collections of real-world grasp attempts~\cite{pinto2016supersizing, pmlr-v87-kalashnikov18a}.

The methods described above see limited applicability in our setting for the following reasons.
1. The use of synthetic data/simulators is limited by the variety and the difficulty in accurately and efficiently simulating granular, deformable foodstuffs at scale.
2. Large-scale data collection is difficult due to concerns of spoilage, damage and hygiene so the training procedure is required to be short.
3. Existing methods focus on grasping a \emph{single solid target object} with fixed shape.
Focus on grasping a single object allows their problem to be cast as a binary success/failure task.
If we were to model our task as a binary classification task, it would be necessary to retrain the network to capture success/failure for grasping every required target weight and would violate one of the key requirements we put forth in Section~\ref{sec:introduction}.

\section{Contributions}
\label{sec:contributions}
In addressing the requirements mentioned in Section~\ref{sec:introduction}, we believe that we have developed a novel system that can reliably grasp a user-specified amount of granular food with a focus on minimizing the data to be collected.
The contributions of this work are as follows:
1) Proposal of a learning-based system for grasping a user-specified mass of various granular food with automatic data labeling.
2) A self-supervised method for modeling its predictive uncertainties.
3) A system that chooses grasps by leveraging model uncertainty to reduce the required amount of training data and to improve accuracy.

\begin{figure}[tb]
	\centering
	\includegraphics[width=0.95\columnwidth]{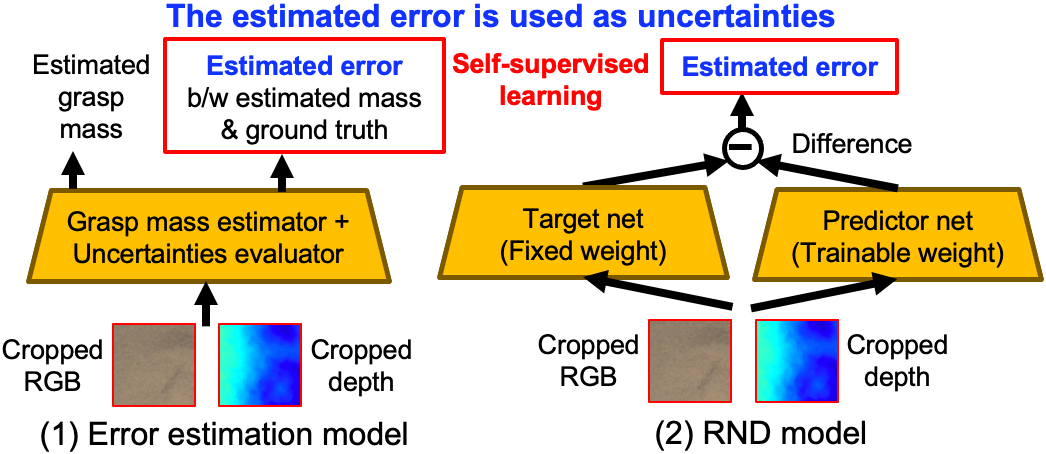}
	\caption{The network models for uncertainties evaluators}
	\label{fig:model}
\end{figure}

\section{Method}
\label{sec:method}
Our proposed method to grasp a user specified target mass of granular food consists of the following steps (see also Fig.~\ref{fig:concept}):
1) Grasp mass predictions are obtained for RGB-D patches across the food tray.
2) The uncertainty in the predicted grasp mass is obtained.
3) Select the point with the lowest uncertainty for the estimated grasp mass among the points whose estimated grasp mass are within $\pm$0.5 g from the target mass.

\subsection{Grasp Mass Estimation from RGB-D}
\label{sec:Grasping Mass Estimation from RGB-D}
Our grasp mass estimation network predicts the mass of food to be grasped for cropped patches of the RGB-D image.
Therefore, if we command the gripper to grasp at a certain patch's location, it is expected that the gripper will grasp the mass that was predicted for that patch.

When the gripper inserts itself into the food during grasping, we find that disturbance is typically localized to the region around the grasp point.
Therefore, we find it sufficient to crop these regions centered around the grasp point. 
Additionally, we process the cropped depth image such that the center of the patch has a value of 0, and the heights of the other points within the patch are relative to it.
This step aims to reduce the diversity of the input data.

The expected grasp mass for a patch $i$ is estimated by the neural network $F(x_{img,i},x_{depth,i};\xi)=y^{\prime}_{mass,i}$ where the input $x_{img,i}$ and $x_{depth,i}$ are the $i$-th cropped RGB image and depth image, respectively.
$\xi$ is a set of parameters to be optimized.
The loss function $L$ is minimized w.r.t $\xi$:
\vspace{-2mm}
\begin{equation}
    \vspace{-2mm}
    \min_{\xi}\frac{1}{n}\sum_{i=1}^{n} L(y_{mass,i}, F(x_{img,i},x_{depth,i};\xi))
    \label{eq:cost_eq}
\end{equation}
where $n$ is the number of cropped images used for the training, and $y_{mass,i}{\in}{R}^{d}$ is the teaching signal for the $i$-th image, i.e., the actual grasp amount.

Possible choices of the loss function $L$, include ones used for regression or classification tasks.
Although the nature of the task is regression because we need to estimate a grasp amount expressed as a continuous value, a classification loss may be used and works well for certain cases (see related studies in Section~\ref{sec:related Grasping}).
This is probably due to the high degree of freedom of multinomial distributions which is optimized by solving the classification task.
In compensation for the high degree of freedom, however, we usually need a large training dataset.
Since we want to work with small training datasets and the output of the network should represent the entire range of possible grasp amounts with a resolution as fine as the used scale (see Section~\ref{sec:Random Grasping for Data Collection}), we chose to optimize a regression loss, MSE in equation~\eqref{eq:cost_eq}.

\subsection{Estimation of Uncertainties}
\label{sec:Estimation of Uncertainties}
In this section, we propose several methods to explicitly capture the uncertainty of the model.
Uncertainty in this context is the likelihood that the actual grasp mass will be different from the predicted mass for the RGB-D input described in Section~\ref{sec:Grasping Mass Estimation from RGB-D}.
Predictions with low uncertainty indicate the model's high confidence that a real grasp's outcome will match its prediction.

As we describe in Section~\ref{sec:related Estimation of Uncertainties}, the uncertainty we estimate here includes both \emph{aleatoric} and \emph{epistemic} uncertainty.
Aleatoric uncertainty is the uncertainty intrinsic to the system such as sensor noise, motor noise, or unpredictable stochastic phenomena.
On the other hand, epistemic uncertainty is uncertainty due to the incompleteness of the estimation method.
In our setting, the shortage of the training data is the primary source of -\emph{epistemic}- uncertainty. 

The uncertainty estimation methods we present below work in tandem with the mass estimation model described in the previous section and can be either stand-alone models or combined with the mass estimation model (see Fig.~\ref{fig:model}).

\subsubsection{Uncertainties by Error Estimation}
\label{sec:Uncertainties by Error Estimation}
Here, we describe a method inspired by literature on improved exploration in Reinforcement Learning~\cite{gehring2013smart}, which we adapt to estimate the uncertainty of the model.
The original method proposes the use of absolute temporal difference errors to identify states in which effects of actions are harder to predict, and use that to encourage/discourage exploration.
This metric is reportedly more robust to noise than the standard deviation.

Given the $i$-th input image, the error estimation (\emph{EE}) model learns to predict the absolute difference, $y^{\prime}_{error, i}$, between the actual grasp amount $y_{mass,i}$ and the estimated grasped amount $y^{\prime}_{mass,i}$ which is estimated by the mass estimation model (see Fig.~\ref{fig:model}~(1)).
The EE model is obtained as the one that minimizes the following loss function:
\vspace{-2mm}
\begin{equation}
    \vspace{-2mm}
    \min_{\xi}\frac{1}{n}\sum_{i=1}^{n} L(|y_{mass,i} - y^{\prime}_{mass,i}|, y^{\prime}_{error,i})
    \label{eq:ee}
\end{equation}
Note here that we do not need any external supervised signal other than $y_{mass,i}$ which we assume available for the training of the mass estimation model.
Thus the training of the EE model is a kind of self-supervised learning.
Since the mass estimation model is trained such that the error is small for the training data, the closer the input data is to the training data, the closer the output of the EE model is to zero.
Furthermore, for the error estimation of the training data, EE model output will reflect the difficulty of the mass estimation because the easier the mass estimation is, the closer the estimated grasp mass will be to the actual grasp amount.
We interpret the estimated error as a proxy measure of model uncertainty.

\subsubsection{Uncertainty estimation inspired by RND}
\label{sec:Uncertainties by Curiosity}
Here we propose to use the idea of Random Network Distillation (\emph{RND})~\cite{burda2018exploration} to estimate the uncertainty of the estimated grasp mass (See Fig.~\ref{fig:model}~(2)).
RND was originally proposed as a technique to explore novel state-action spaces for better exploration in reinforcement learning.
When exploring, it is necessary to determine whether something has already been experienced, or has not yet been experienced, and RND helps determine this.
In RND, two networks $(f, f^{\prime})$ are used.
One is the \emph{target network} $f$ which maps an observation to an embedding $f: O \mapsto {R}^{k}$, and the other is the \emph{predictor network} $f^{\prime}$ for $f^{\prime}: O \mapsto {R}^{k}$.
The target network produces teaching signals for the training of the predictor network.
In practice, it is represented by a random neural network whose weights are randomly initialized and kept fixed.
Then, the predictor network is trained by gradient descent to minimize the average MSE$\frac{1}{n}\sum_{i=1}^n||f^{\prime}(x_{img,i}, x_{depth,i};\theta) - f(x_{img,i}, x_{depth,i})||^{2}$ with respect to its parameter $\theta$ in our case.
Note here that $\{ x_{img,i}, x_{depth,i} \}$ are exactly the same images used for the training of the mass estimation model, but the teaching signal is given by the target network.
Therefore, when $f$ and $f^{\prime}$ are given inputs outside of the training set, this difference is expected to be large.
We adopt RND's property to model uncertainty, i.e., the closer the input is to the training data, the lower the uncertainty is for the output of the mass estimation model.
This method requires two networks $f$ and $f^{\prime}$ in addition to the mass estimation model but doesn't affect its learning.

\subsection{Grasp Point Selection}
\label{sec:Grasp Point Selection}
In this section, we describe how we select the grasp point from the food tray.
We start with the case of a model using ordinary mass estimation without uncertainty.
Patches centered around each candidate grasp point are cropped from the RGB-D data and fed into the mass estimation model.
The patch with the predicted mass closest to the target grasp mass is selected for execution.
This method works well if there is enough data to train the mass estimation model, as the difference between the estimated and actual grasped masses will be small.
If the training data is not sufficient, the difference between the actual grasp and the estimate is sometimes large because the grasp estimation model is not sufficiently accurate.
As our work deals with granular foods, a large amount of data is needed to capture the immense variety of their possible configurations.
This, coupled with the intrinsic randomness in the grasp process makes it unlikely to ever obtain a perfect model.
Therefore, we propose to augment the method with uncertainty models as described in the previous section.

We use a two-stage grasp selection criterion where the method first filters the points for which the estimated masses are within an acceptable region, which is specified to be within $\pm$0.5 g from the target mass, which corresponds to the resolution of the scale we used for creating the teaching signal.
Then, we select the ones with the lowest uncertainty amongst them to maximize the probability to grasp the food of the estimated grasp amount. 
We expect that this two stage process with the inclusion of uncertainty in the selection process achieves significant improvements over the ones that didn't, particularly when the size of the training samples is small.
Using only the mass estimation model to pick a grasp point is equivalent to choosing amongst the qualifying candidate uncertainties uniformly at random.  Choosing the candidate with least uncertainty is expected to be as good as, or better than uniformly sampling them.

\section{Experiment Setup for Food Picking}
\label{sec:experiments}
The goal of the experiments is to verify and compare several models that can grasp a target amount of granular food, which make use of uncertainties to improve performance.

\begin{figure}[tb]
	\centering
	\includegraphics[width=0.90\columnwidth]{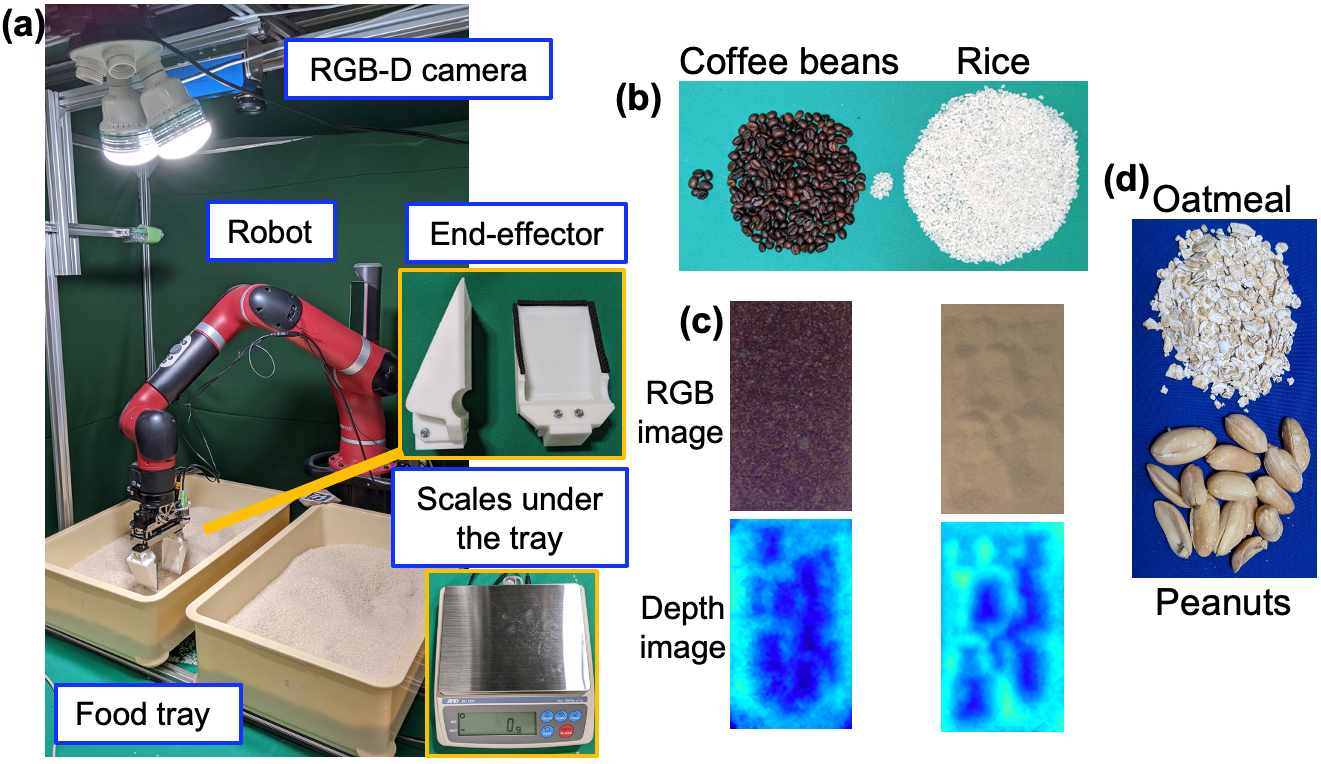}
	\caption{Experiment setup: (a) robot setup used for the experiment, (b) target object images; coffee beans with 1 g and 22 g (left), and rice with 1 g and 60 g (right), (c) examples of RGB and depth images of coffee beans (left) and rice (right), (d) oatmeal (top) and peanuts (bottom)
	}
	\label{fig:setup}
\end{figure}
\subsection{Robot Setup for Data Collection System}
\label{sec:Robot Setup for Data Collections System}
Our robotic system, shown in~\Cref{fig:setup}~(a), consists of a Sawyer 7-DOF robotic arm equipped with a self-designed two-fingered parallel gripper driven by a servo motor (Dynamixel XM430-W350-R).
A sponge is attached to the edge of the end-effector since a gap occurs when the food is pinched, and this sponge prevents the food from spilling out (See black lining around the end-effector in Fig.~\ref{fig:setup}~(a)).
Furthermore, we use an Ensenso N35 stereo camera in combination with an IDS uEye RGB camera to overlook the workspace of the robot arm and use them to retrieve registered point clouds of the scene.
We use two food trays and the robot continuously picks from the one to the other and vice versa until a user-specified number of iterations.
The inner dimensions of the food tray are $603 mm \times 377 mm \times 145 mm$ which is a commonly used tray size in the food industry.
Two EW-12Ki scales are placed under each food tray, and the amount grasped by the robot is calculated from the masses of the trays after the robot picks from the one and after it places in the other.
The Sawyer, gripper, RGB-D sensor, and scales are connected to a PC running Ubuntu 16.04 with ROS Kinetic.
We target coffee beans and uncooked rice which have different particle sizes, surface friction and densities (See Fig.~\ref{fig:setup}~(b)).
Additionally we show that our methods work well for two other foods, namely oatmeal and peanuts (See Fig.~\ref{fig:setup}~(d))
We primarily focus on the grasping of generic dry, cohesion-less granular foodstuffs.
Non-dry viscous foods or ones that are prone to tangling are outside the scope of our work and would likely require hardware modifications and additional manipulation such as pre-grasping as seen in~\cite{ray2020robotic}.
We find our system works reasonably well on most granular materials that satisfy the above criterion and have enough mass to measure well on our scale which has a resolution of 1 g.
\subsection{Data Collection Process}
\label{sec:Data Collection Process}
We built a data collection system to collect data indefinitely, without the need for human supervision or labeling effort.
Algorithm~\ref{alg:datacollection} details the data collection process.

We choose the home position such that the robot is out of view of the RGB-D camera.
To decide which tray to pick from and which to place in, we weigh the trays and choose the heaviest one to pick from first.
Once the mass of the pick tray becomes lower than a certain threshold, the robot switches direction, so the current pick tray will become the place tray and vice versa.
After deciding the pick/place trays, the system captures an RGB-D image and fills in all NaN values, as seen in lines 6-7 of Algorithm 1.

In line 8 of Algorithm 1, we process the RGB-D images in the following ways.
From the entire RGB-D image, we obtain cropped patches of 150x150 pixels with the grasp point as the center of the patch.
1 pixel corresponds to roughly 1 mm of the tray.
The depth information is used to compute the relative heights from the food surface's height at the grasping point w.r.t. the height at the center of the cropped patch.
Note that this is an arbitrary choice to normalize input height.
Then it was converted to RGB by OpenCV's pseudo-coloring API, applyColorMap()~\cite{bradski2008learning}.

Note that in line 10, the pick and place poses are randomly selected within the tray areas, which are located through Aruco markers at line 9~\cite{garrido2016generation, romero2018speeded}.
Only the $x$ and $y$ coordinates are randomized though, while the pick height is always set to be 2 cm below the surface of the food.
By fixing the grasp height, we ensure a more consistent and diverse dataset due to the reduction in exploration of the height parameter, and the constantly decreasing height of the food as the bin empties, leading to a variety of grasped volumes and surface conditions.
Though outside the scope of this work, when volume of food becomes too small to grasp well, simple motions can be executed to gather the remaining food into piles.

This picking system uses two food trays, and food grasped from one tray is placed in the other.
Therefore, the total amount of food in the tray being picked from continues to decrease as the other increases. The surface of the food in both trays continue to evolve significantly and constantly (See Fig.~\ref{fig:setup}~(c)).
This allows us to capture diverse surface configurations that may arise in a real factory setting.

The whole pick-and-place process, including data storage, takes about 12.5 sec per cycle (See attached video).
During inference, it takes an additional 1.5 sec, leading to a total of 14.0 sec per cycle.
For the training data, we collected 1300 data points for rice and coffee, respectively.
Out of 1300 samples collected, 1000 were used for training, and 300 were used to evaluate the network model.
During training of the network(s), we improve generalization by exploiting the gripper symmetry and apply vertical and horizontal flipping of input RGB-D independently with a probability of 0.5.

\algdef{SE}[DOWHILE]{Do}{doWhile}{\algorithmicdo}[1]{\algorithmicwhile\ #1}%
\begin{algorithm}[tb]
    \caption{Data Collection Process}
    \label{alg:datacollection}
    \begingroup
        \scalefont{0.75}
        \begin{algorithmic}[1]
        \For{$\rm{\textsc{iter}}=1$ to $N$}
        	\State $q_{angles} \gets \textsc{home}$ \Comment{Move out of view.}
        	\State $tray_{pick} \gets
        	\textsc{SelectTray()}$
        	\Comment{Select the pick tray.}
        	\If{$mass_{tray} < \epsilon$}
        	    \textsc{SwitchDirection()}
            \EndIf
            \State $RGBD \gets \textsc{CaptureImage\&Depth()}$
            
            \While{\textsc{Depth.ContainsNaN()}}
                \State \textsc{RunningAverage()} 
            \EndWhile
            \State \textsc{process(RGBD)}
            \State $P_{tray_{pick}}, P_{tray_{place}} \gets \textsc{FindTrayLocations()}$
            
            \State $X, Y \gets \textsc{GenerateRandomXYCoordinates()}$ 
            \State $\textsc{Pick(X, Y, Z = 0.02$ m$)}$ \Comment{Pick at fixed height from the food surface}
            \State $\textsc{Place()}$ \Comment{Drop in $tray_{place}$}
        	\State \textsc{MeasureScales()} \Comment{Measure picked amount}
        \EndFor
        \end{algorithmic}
    \endgroup
\end{algorithm}

\subsection{Deep Learning \& Training \& Inference}
\label{sec:Deep Learning}
We trained the networks on a machine equipped with 256\,GB RAM, an Intel Xeon E5-2667v4 CPU, and Tesla P100-PCIE with 12GB.
Training for each model (initialized randomly for each food) varied from 10 to 70 minutes for difference dataset size.
Our experiments for inference with trained models were performed on a machine equipped with 31.3\,GB RAM, an Intel Core i7-7700K CPU, and GeForce GTX 1080 Ti.
Inference time was about 1.5 sec.~\footnote{Details of architectures for mass estimation and uncertainty prediction are shown in the video, but note that other architectures can also be used.}
\begin{figure}[tb]
	\centering
	\includegraphics[width=1.0\columnwidth]{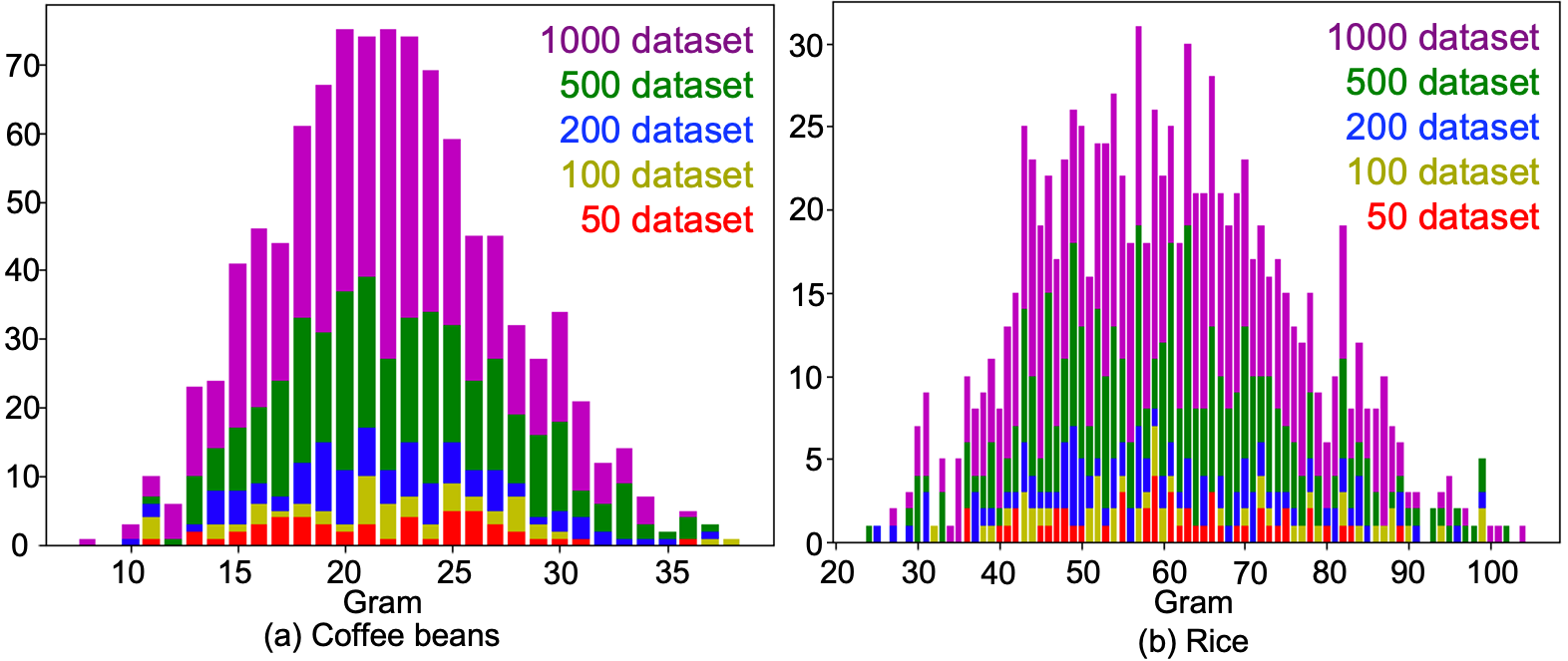}
	\caption{Histogram of (a) coffee beans and (b) rice for 50, 100, 200, 500, and 1000 times grasps randomly for training data.}
	\label{fig:histogram}
\end{figure}

\begin{table*}[tb]
    \centering
    \caption{Success rates for coffee beans and rice. Numbers in bold represent better success rates than \emph{Baseline}.}
    \label{tab:coffee beans and rice success rate}
    \begingroup
    \scalefont{0.85}
        \begin{tabular}{c|c||c c | c c | c c||c c | c c | c c}
        \hline
        \multirow{4}{*}{\shortstack{Dataset sizes\\for training}} & \multirow{4}{*}{Model} & \multicolumn{6}{c||}{Coffee beans} & \multicolumn{6}{c}{Rice} \\
        \cline{3-14}
        && \multicolumn{2}{c|}{17 g} & \multicolumn{2}{c|}{22 g} & \multicolumn{2}{c||}{27 g}
         & \multicolumn{2}{c|}{45 g} & \multicolumn{2}{c|}{60 g} & \multicolumn{2}{c}{75 g} \\
        \cline{3-14}
        && \multirow{2}{*}{\shortstack{$5\%$\\$(1 g)$}} & \multirow{2}{*}{\shortstack{$10\%$\\$(2 g)$}} &
           \multirow{2}{*}{\shortstack{$5\%$\\$(2 g)$}} & \multirow{2}{*}{\shortstack{$10\%$\\$(3 g)$}} &
           \multirow{2}{*}{\shortstack{$5\%$\\$(2 g)$}} & \multirow{2}{*}{\shortstack{$10\%$\\$(3 g)$}}
         & \multirow{2}{*}{\shortstack{$5\%$\\$(3 g)$}} & \multirow{2}{*}{\shortstack{$10\%$\\$(5 g)$}} &
           \multirow{2}{*}{\shortstack{$5\%$\\$(3 g)$}} & \multirow{2}{*}{\shortstack{$10\%$\\$(6 g)$}} & 
           \multirow{2}{*}{\shortstack{$5\%$\\$(4 g)$}} & \multirow{2}{*}{\shortstack{$10\%$\\$(8 g)$}} \\
        &&&&&&&&&&&&&\\
        \hline\hline
            \multirow{4}{*}{50}
                & Random   & 0.22 & 0.32 & 0.22 & 0.38 & 0.32 & 0.36 
                           & 0.16 & 0.24 & 0.24 & 0.40 & 0.16 & 0.28 \\
                & Baseline & 0.34 & 0.52 & 0.56 & 0.78 & 0.42 & 0.52 
                           & 0.24 & 0.32 & 0.32 & 0.58 & 0.34 & 0.67 \\
                & EE       & \bf{0.52} & \bf{0.82} & \bf{0.76} & \bf{0.94} & \bf{0.74} & \bf{0.82} 
                           & \bf{0.32} & \bf{0.49} & \bf{0.54} & \bf{0.79} & \bf{0.52} & \bf{0.73} \\
                & RND      & \bf{0.46} & \bf{0.70} & \bf{0.64} & \bf{0.82} & \bf{0.52} & \bf{0.64} 
                           & \bf{0.24} & \bf{0.40} & \bf{0.33} & \bf{0.65} & \bf{0.45} & \bf{0.77} \\
            \hline
            \multirow{4}{*}{100}
                & Random   & 0.17 & 0.25 & 0.29 & 0.43 & 0.31 & 0.36
                           & 0.15 & 0.20 & 0.22 & 0.33 & 0.15 & 0.25 \\ 
                & Baseline & 0.48 & 0.64 & 0.60 & 0.76 & 0.60 & 0.84 
                           & 0.53 & 0.81 & 0.58 & 0.83 & 0.36 & 0.75 \\
                & EE       & \bf{0.54} & \bf{0.84} & \bf{0.80} & \bf{0.92} & \bf{0.80} & \bf{0.88} 
                           & \bf{0.64} & \bf{0.82} & \bf{0.66} & \bf{0.96} & \bf{0.38} & \bf{0.76} \\
                & RND      & \bf{0.56} & \bf{0.76} & \bf{0.64} & \bf{0.86} & \bf{0.68} & \bf{0.86} 
                           & \bf{0.62} & \bf{0.82} & \bf{0.59} &     0.82  & \bf{0.55} & 0.73 \\
            \hline
            \multirow{4}{*}{200}
                & Random   & 0.14 & 0.255 & 0.315 & 0.465 & 0.25  & 0.32
                           & 0.14 & 0.22  & 0.19  & 0.295 & 0.135 & 0.255 \\
                & Baseline & 0.42 & 0.68  & 0.56  & 0.82  & 0.68  & 0.82 
                           & 0.60 & 0.78  & 0.60  & 0.88  & 0.46  & 0.76 \\
                & EE       & \bf{0.62} & \bf{0.80}  & \bf{0.80}  & \bf{0.94}  & \bf{0.76} & \bf{0.84} 
                           & \bf{0.64} & \bf{0.80}  &     0.56   & \bf{0.90}  & \bf{0.64} & \bf{0.90} \\
                & RND      & \bf{0.66} & \bf{0.84}  & \bf{0.74}  & \bf{0.82}  & \bf{0.76} & \bf{0.90} 
                           & \bf{0.64} & \bf{0.88}  & \bf{0.70}  & \bf{0.92}  & \bf{0.62} & \bf{0.94} \\

            \hline
            \multirow{4}{*}{500}
                & Random   & 0.154 & 0.25  & 0.34 & 0.466 & 0.236 & 0.34
                           & 0.138 & 0.214 & 0.19 & 0.308 & 0.136 & 0.26 \\ 
                & Baseline & 0.50  & 0.86  & 0.74 & 0.84  & 0.84  & 0.92 
                           & 0.56  & 0.78  & 0.54 & 0.84  & 0.70  & 0.94 \\
                & EE       & \bf{0.50}  &     0.84  & \bf{0.80} & \bf{0.94} & \bf{0.92} & \bf{0.96} 
                           & \bf{0.68}  & \bf{0.92} & \bf{0.58} & \bf{0.92} &     0.54  & \bf{0.94} \\
                & RND      & \bf{0.58}   &     0.80  & \bf{0.90} & \bf{0.96} &     0.80  & \bf{0.92} 
                           & \bf{0.58}  & \bf{0.80} & \bf{0.54} & \bf{0.86} & \bf{0.70}  & \bf{0.96} \\

            \hline
            \multirow{4}{*}{1000}
                & Random   & 0.151 & 0.259 & 0.367 & 0.493 & 0.208 & 0.311
                           & 0.144 & 0.216 & 0.17  & 0.307 & 0.133 & 0.26 \\ 
                & Baseline & 0.56  & 0.70  & 0.86  & 0.98  & 0.86  & 0.92 
                           & 0.66  & 0.92  & 0.60  & 0.98  & 0.62  & 1.00 \\
                & EE       & \bf{0.64}  & \bf{0.88}  &     0.80   &     0.90  &     0.72   &     0.88 
                           & \bf{0.66}  &     0.88   &     0.56   &     0.86  &     0.54   &     0.92 \\
                & RND      & \bf{0.66}  & \bf{0.80}  & \bf{0.90}  & \bf{0.98} & \bf{0.86}  & \bf{0.96} 
                           &     0.64   & \bf{0.92}  & \bf{0.72}  & \bf{1.00} & \bf{0.68}  &     0.94 \\
            \hline
        \end{tabular}
    \endgroup
    \vspace{-5.0mm}
\end{table*}

\begin{table}[tb]
    \caption{Success rates for oatmeal and peanuts with 50 dataset.}
    \label{tab:oatmeal and peanuts success rate}
    \centering
    \begingroup
    \scalefont{0.85}
        \begin{tabular}{ c ||c c || c c}
        \hline
        \multirow{3}{*}{Model} & \multicolumn{2}{c||}{20 g (Oatmeal)} & \multicolumn{2}{c}{30 g (Peanuts)}\\
        \cline{2-5}
        & \multirow{2}{*}{\shortstack{$5\%$\\$(1 g)$}} & \multirow{2}{*}{\shortstack{$10\%$\\$(2 g)$}}
        & \multirow{2}{*}{\shortstack{$5\%$\\$(2 g)$}} & \multirow{2}{*}{\shortstack{$10\%$\\$(3 g)$}} \\
        &&&&\\
        \hline\hline
                Random     & 0.28 & 0.34 
                           & 0.20 & 0.30     \\
                Baseline & 0.36 & 0.54     
                           & 0.31 & 0.43 \\
                EE       & \bf{0.37} & \bf{0.62}  
                           & \bf{0.44} & \bf{0.61}  \\
                RND      & \bf{0.46} & \bf{0.65}  
                           & \bf{0.36} & \bf{0.48}  \\
            \hline
        \end{tabular}
    \endgroup
\end{table}
\section{Experiment Results}
\label{sec:results}
In our experiments, we compare the following methods: A) \emph{Random:} Uniformly sample grasp points within the tray (this is the data collection method)
B) \emph{Baseline:} Evaluate 900 equidistant grasp points from a 45$\times$20 grid with about 1 cm grid size and select the point where mass estimation model's output is closest to the target mass.
C) \emph{EE:} \emph{Baseline} with uncertainty estimation by the proposed error estimation model and, D) \emph{RND:} same as C but using the RND model.

\subsection{Random Grasping for Data Collection}
\label{sec:Random Grasping for Data Collection}
Figure~\ref{fig:histogram} shows the frequency with which a particular mass of coffee beans or rice was grasped.
We show frequencies for training datasets of sizes 50, 100, 200, 500, and 1000. 
These datasets were also used to train our neural networks.

For both coffee beans and rice, the data increasingly looks normally distributed for sizes 200 onwards.
We also note that the variance of coffee beans is smaller than that of rice due to rice's higher density leading to more varied outcomes.

For the smaller dataset sizes of 50 and 100, we note the data frequency is a lot more discontinuous and sparse.
Some values like 60 g of rice are notably absent in the dataset of size 50, despite being one of the evaluated targets in the next section.
Methods trained with classification would not interpolate well and compensate for these missing values as they have no concept of their labels' continuity.

\subsection{Target-mass Grasping with Uncertainties}
\label{sec:Target-mass Grasping with Deep Learning}
We collect datasets as described in Section~\ref{sec:Random Grasping for Data Collection} and partition them into datasets of size 50, 100, 200, 500, and 1000.  
We train the methods described in Section~\ref{sec:method} for each of these datasets and evaluate them by attempting to grasp three target masses each. 
Using the training set of size 1000, we calculate the \emph{mean} and standard deviation (\emph{std}) of the grasped masses and use them to set the grasp targets \emph{(mean-std, mean, and mean+std)}.
Coffee beans were grasped at targets of (17 g, 22 g, and 27 g), and rice at (45 g, 60 g, and 75 g), respectively. 
We define a \emph{successful} grasp as one which is within an acceptable tolerance range of the required target (expressed in \%).
More specifically, we count a grasp as a success when $\frac{| predicted\_mass - target\_mass|}{target\_mass} \leq tol $. 
We present success rates in Table~\ref{tab:coffee beans and rice success rate} for two tolerance values- 5\% and 10\%, which are the accepted food industry standards for human picks and robot picks, respectively.

For each method, we attempt to grasp every target mass 50 times (with rice trained with datasets of size 50 and 100 being the exceptions with 100 attempts required to stabilize results).
To select the grasping point, 900 points are selected via grid search over the food tray, with a resolution of roughly 1cm.
Among these 900 points evaluated by the model, there are typically more than 10 points within the acceptable range of $\pm$0.5 g (See Fig.~\ref{fig:histogram}).

From Table~\ref{tab:coffee beans and rice success rate}, we see that methods that incorporate uncertainty in the grasp selection process, namely \emph{EE} and \emph{RND}, tend to outperform the \emph{baseline}, especially for the small dataset sizes.
We highlight these results in bold.
For datasets of size 50-200, we see both \emph{EE} and \emph{RND} outperform the baseline in almost every instance, with the difference being particularly stark for size 50.
Table~\ref{tab:coffee beans and rice success rate} also shows that for coffee beans, the performances of \emph{EE} and \emph{RND} trained on a dataset of size 50 is similar to that of the baseline trained on datasets of size 200 or larger.
Although the experimental results for rice do not show as remarkable a difference as coffee beans, the use of uncertainty still shows a marked performance improvement.
For rice we observe that \emph{EE} and \emph{RND} reach performances similar to the baseline, but consistently requiring 50-100 less training samples.

The \emph{random} method, which we present as the null hypothesis, is always outperformed by other methods.
We note that as the size of the dataset increases, \emph{baseline} catches up to or sometimes exceeds the performance of the methods we present.
This behavior is as expected, as the increasing training dataset sizes decrease the epistemic uncertainty in the model's prediction as it is better able to capture the data generating process.
Incorporating these uncertainty predictions in the grasp selection process no longer provides pertinent information that isn't already captured in the baseline's maximum likelihood estimate.

The lack of significant improvement for large datasets suggests methods like \emph{EE} and \emph{RND} predominantly capture epistemic uncertainty, which is considered reducible by the gathering of additional data.
However, since data comes at a premium for applications like ours, it is essential to minimize it as much as possible.
Incorporating uncertainty into the decision process provides a practical, easy-to-implement method to learn with significantly fewer data samples.

We study the effect of varying data-size and target masses on coffee beans and rice and confirm the possibility of small-dataset training.
Furthermore, we test its veracity via additional experiments conducted on specific target weights for oatmeal and peanuts.
Oatmeal and peanuts have yet again different dynamics compared to coffee beans and rice, but Table~\ref{tab:oatmeal and peanuts success rate} shows that our methods still work robustly while using the same gripper.
Oatmeal is light and planar and gets compressed when grasped, while peanuts break into halves when grasped too tightly.
These differences heavily influence the nature of the collected data, but despite of this our methods were able to capture the dynamics of the different granular foods well using few training data.

\section{Conclusion}
\label{sec:conclusion}
In this paper, we introduce methods to capture a model's uncertainty through self-supervised learning and use it in grasp selection to obtain competitive performance using remarkably smaller data sizes.
Additionally, we have developed an autonomous data collection system.
To capture the uncertainty, we adopt \emph{EE} and \emph{RND} models, and apply them to our grasp mass estimation models for granular foods.
We show empirically that a simple modification to a baseline grasp selection heuristic that chooses the least uncertain prediction among the grasp candidates results in significant performance improvements.
This modification allows for success rates similar to the baseline, but requires less training data compared to the baseline.

\section*{ACKNOWLEDGMENT}\small
The authors would like to thank Dr. Naoki Fukaya for creating the grippers used in our experiments.
\bibliographystyle{IEEEtran} 
\bibliography{IEEEabrv,bibliography}
\end{document}